\documentclass[conference]{IEEEconf}
\IEEEoverridecommandlockouts

\usepackage{cite}
\usepackage{amsmath,amssymb,amsfonts}
\usepackage[ruled,vlined]{algorithm2e}
\usepackage{graphicx}
\usepackage{textcomp}
\usepackage{xcolor}
\usepackage{booktabs}
\usepackage{multirow}
\usepackage[colorlinks]{hyperref}
\usepackage{pifont}
\usepackage[normalem]{ulem}

\def\BibTeX{{\rm B\kern-.05em{\sc i\kern-.025em b}\kern-.08em
    T\kern-.1667em\lower.7ex\hbox{E}\kern-.125emX}}
\begin{document}

\newcommand{\blue}[1]{
{\color{blue}{#1}}
}

\newcommand{\red}[1]{
{\color{red}{#1}}
}

\newcommand{\green}[1]{
{\color{green}{#1}}
}


\title{\LARGE \bf 

STRIDE: Structured Lagrangian and Stochastic Residual Dynamics via Flow Matching

}


\author{
Prakrut Kotecha$^{1}$, Ganga Nair B$^{1}$, Shishir Kolathaya$^{2}$
\thanks{$^{1}$G. Nair, P. Kotecha, are with the Robert Bosch Center for Cyber-Physical Systems, Indian Institute of Science, Bengaluru.}%
\thanks{$^{2}$S. Kolathaya is with the Robert Bosch Center for Cyber-Physical Systems and the Department of Computer Science \& Automation, Indian Institute of Science, Bengaluru.}
\thanks{Email: \href{mailto:stochlab@iisc.ac.in}{stochlab@iisc.ac.in}, Website: \href{https://prakrutk.github.io/STRIDE/}{link}}%
}
\maketitle

\begin{abstract}

Robotic systems operating in unstructured environments must operate under significant uncertainty arising from intermittent contacts, frictional variability, and unmodeled compliance. While recent model-free approaches have demonstrated impressive performance, many deployment settings still require predictive models that support planning, constraint handling, and online adaptation. Analytical rigid-body models provide strong physical structure but often fail to capture complex interaction effects, whereas purely data-driven models may violate physical consistency, exhibit data bias, and accumulate long-horizon drift. In this work, we propose STRIDE, a dynamics learning framework that explicitly separates conservative rigid-body mechanics from uncertain, effectively stochastic non-conservative interaction effects. The structured component is modeled using a Lagrangian Neural Network (LNN) to preserve energy-consistent inertial dynamics, while residual interaction forces are represented using Conditional Flow Matching (CFM) to capture multi-modal interaction phenomena. The two components are trained jointly end-to-end, enabling the model to retain physical structure while representing complex stochastic behavior. We evaluate STRIDE on systems of increasing complexity, including a pendulum, the Unitree Go1 quadruped, and the Unitree G1 humanoid. Results show 20\% reduction in long-horizon prediction error and 30\% reduction in contact force prediction error compared to deterministic residual baselines, supporting more reliable model-based control in uncertain robotic environments.


\end{abstract}
\section{Introduction}

Robotic systems are increasingly expected to operate outside controlled laboratory settings and within unstructured, dynamic environments. In such conditions, robots must contend with significant uncertainty arising from intermittent contacts, frictional variability, unmodeled compliance, and actuator nonlinearities \cite{hu2019chainqueen}. These effects are particularly pronounced in legged and humanoid systems, where small disturbances during contact can lead to large deviations in system behavior.

Recent advances in deep reinforcement learning have demonstrated that model-free policies can achieve highly agile behaviors directly from observation, especially when combined with large-scale training and domain randomization \cite{tan2018sim, agilejump}. However, purely reactive policies provide limited scope for reasoning about future state evolution, enforcing hard safety constraints, or adapting behavior online when encountering previously unseen conditions. Many real-world deployment scenarios therefore continue to benefit from predictive models that enable planning, constraint handling, and closed-loop adaptation \cite{nagabandi2018neural, chua2018deep}.

Model-based control frameworks offer these capabilities by explicitly leveraging system dynamics. When a sufficiently accurate model is available, control methods such as Model Predictive Control (MPC) and trajectory optimization can incorporate task objectives at runtime, enforce state and input constraints, and locally replan under distribution shift. In practice, however, obtaining dynamics models that are both physically consistent and expressive enough to capture real-world interaction uncertainty remains a central challenge.

This requirement becomes particularly critical in high-dimensional robotic platforms such as quadrupeds and humanoids, where planning and control operate under tight stability margins and frequent contact events \cite{raibert1986legged,herzog2016momentum,werling2021fast}. In these regimes, small modeling errors can compound rapidly, degrading planning performance or even leading to unstable motion \cite{kotecha2025investigating, schulze2025floating}.

\begin{figure}[t]
    \centering
    \includegraphics[width=\linewidth]{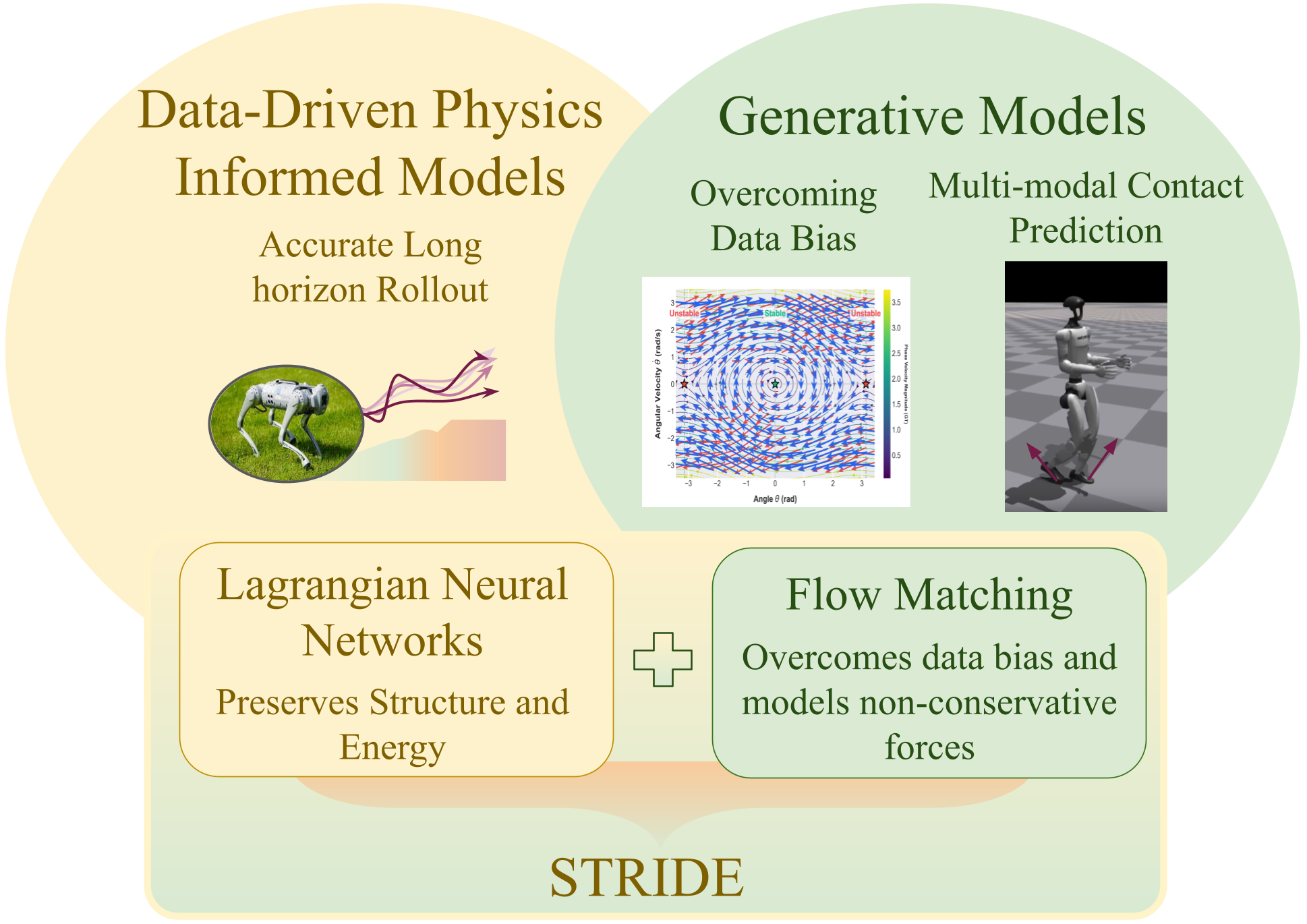}
    \caption{Conceptual overview of STRIDE. The proposed framework combines a structured Lagrangian prior with a stochastic residual to capture interaction uncertainty while preserving physical consistency. \vspace{-20pt}}
    \label{fig:concept_image}
\end{figure}

Analytical rigid-body models provide strong physical structure grounded in first-principles and remain central to many control and planning pipelines \cite{siciliano2010robotics,featherstone2014rigid,todorov2012mujoco}. When accurate parameters are available, such models preserve important properties including inertia coupling and momentum consistency. However, deriving high-fidelity analytical models for complex robotic systems is often difficult and computationally expensive, and their accuracy deteriorates in the presence of frictional changes, impacts, and unmodeled compliance.

Purely data-driven approaches have emerged as an alternative by learning system dynamics directly from trajectory data \cite{deisenroth2011pilco,williams2017information}. While expressive, such models typically lack physical inductive bias, require large datasets for generalization, and may violate fundamental physical constraints, leading to energy inconsistency and compounding long-horizon prediction errors \cite{gruverdeconstructing}. 

To bridge analytical and learning-based paradigms, physics-informed neural architectures embed structural priors directly into learned models. In particular, Lagrangian Neural Networks (LNNs) parameterize system dynamics through learned kinetic and potential energy functions, enforcing the Euler-Lagrange equations by construction \cite{lutter2019deep,cranmer2020lagrangian}. Related formulations, including Hamiltonian Neural Networks and Neural ODE models, similarly incorporate physics-inspired inductive biases \cite{greydanus2019hamiltonian,chen2018neuralode}. These approaches improve physical consistency and sample efficiency but typically represent non-conservative effects such as friction and contact impulses using deterministic residual terms \cite{zhong2021extending,schulze2025context,kotecha2025investigating}. However,  due to unresolved degrees of freedom and partial observability, a purely deterministic representation of non-conservative forces fails to capture the underlying stochasticity inherent in such complex interactions.

Recent efforts to capture this stochasticity have explored fully generative transition models, including diffusion and score-based approaches \cite{pan2024model,ho2020denoising,xu2025dynamics}. While expressive, these models often lack the structural decompositions required for linearization-based planning, and embedding iterative generation within real-time control loops introduces prohibitive computational overhead. To retain expressive power while supporting predictable inference, Conditional Flow Matching (CFM) offers an attractive alternative: by learning continuous transport maps, CFM enables direct, efficient sampling from complex conditional distributions without the multi-step denoising required by diffusion models \cite{lipmanflow}.

Motivated by these observations, we propose \textbf{STRIDE}, a dynamics learning framework that separates conservative rigid-body mechanics from stochastic non-conservative effects (Fig.~\ref{fig:concept_image}). The structured component is modeled using an LNN to preserve the inductive biases of analytical mechanics, while residual interaction forces are modeled using CFM to capture multi-modal variability efficiently. By doing so, STRIDE preserves physical consistency while reducing the modeling burden on the generative component.

We evaluate STRIDE on the Unitree Go1 quadruped and the Unitree G1 humanoid. Empirically, the proposed approach improves long-horizon prediction error and more faithfully captures contact transitions compared to deterministic residual baselines. We further demonstrate the effectiveness of STRIDE within model-based control pipelines through both simulation studies on Unitree Go1 and G1 and real-world experiments on the Unitree Go1, highlighting its practical deployability. We also perform an additional experiment on pendulum to show its ability to preserve topology of dynamics of the system.

\section{Problem Formulation}

\subsection{Mechanical System Model}

We consider a robotic system defined by generalized coordinates $\mathbf{q} \in \mathbb{R}^n$ and generalized velocities $\dot{\mathbf{q}} \in \mathbb{R}^n$. In an ideal, undisturbed environment, the system is to evolve according to classical Lagrangian mechanics. The scalar Lagrangian $(\mathcal{L})$ is defined as:
\begin{equation}
\mathcal{L}(\mathbf{q}, \dot{\mathbf{q}}) = 
\mathcal{T}(\mathbf{q}, \dot{\mathbf{q}}) - 
\mathcal{V}(\mathbf{q}),
\label{eq:lagrangian}
\end{equation}
where $\mathcal{T}$ and $\mathcal{V}$ denote the kinetic and potential energy, respectively. The evolution of the system follows the Euler-Lagrange equations:
\begin{equation}
\frac{d}{dt} \left( \frac{\partial \mathcal{L}}{\partial \dot{\mathbf{q}}} \right)
-
\frac{\partial \mathcal{L}}{\partial \mathbf{q}}
=
\boldsymbol{\tau} + \mathbf{F}_{\text{ext}},
\label{eq:euler_lagrange}
\end{equation}
where $\boldsymbol{\tau} \in \mathbb{R}^n$ represents the known control inputs and $\mathbf{F}_{\text{ext}}$ represents the sum of all external, non-conservative, and unmodeled forces. By expanding \eqref{eq:euler_lagrange}, we can arrive at the standard rigid-body dynamics form:
\begin{equation}
\mathbf{M}(\mathbf{q}) \ddot{\mathbf{q}} 
+ \mathbf{C}(\mathbf{q}, \dot{\mathbf{q}}) \dot{\mathbf{q}} 
+ \mathbf{g}(\mathbf{q})
=
\boldsymbol{\tau} + \mathbf{F}_{\text{ext}},
\label{eq:rbd}
\end{equation}
where $\mathbf{M}(\mathbf{q})$ is the positive-definite mass matrix, $\mathbf{C}(\mathbf{q}, \dot{\mathbf{q}})$ captures Coriolis and centrifugal effects, and $\mathbf{g}(\mathbf{q})$ represents gravitational forces.




\subsection{Structured vs. Residual Dynamics}

Equation~\eqref{eq:rbd} motivates a useful modeling decomposition of robot dynamics. The rigid-body terms capture the dominant, geometry-dependent mechanics arising from the robot’s morphology (e.g., mass distribution and inertial coupling), while the external force term $\mathbf{F}_{\text{ext}}$ aggregates non-conservative and environment-dependent effects.

In practice, while the rigid-body component is smooth and amenable to physics-informed parameterization, $\mathbf{F}_{\text{ext}}$ acts as an aggregation of non-conservative, state-dependent effects (e.g., friction and impacts) that often exhibit multi-modal behavior near contact transitions.


Motivated by this observation, STRIDE models dynamics as: the dominant rigid-body dynamics using a structured Lagrangian prior, while the remaining interaction effects are represented using a stochastic residual model. This decomposition is not enforced strictly but serves as a practical inductive bias that improves learning efficiency and long-horizon predictive stability in complex robotic settings.

\subsection{Learning Objective}

Our goal is to learn a predictive model $\ddot{\mathbf{q}} = f_\theta(\mathbf{q}, \dot{\mathbf{q}}, \boldsymbol{\tau})$ that does not sacrifice physical consistency for expressive power. We propose a structural decomposition of the predicted acceleration:
\begin{equation}
\ddot{\mathbf{q}}_{\text{pred}} = f_{\text{LNN}}(\mathbf{q}, \dot{\mathbf{q}}, \boldsymbol{\tau}) + \mathbf{M}^{-1}(\mathbf{q})\boldsymbol{\epsilon}_{\text{CFM}}(\mathbf{q}, \dot{\mathbf{q}}, \boldsymbol{\tau}),
\end{equation}
where:
\begin{enumerate}
    \item $f_{\text{LNN}}$ is a deterministic \textbf{LNN} that learns the conservative prior by parameterizing $\mathbf{M}(\mathbf{q})$ and $\mathcal{V}(\mathbf{q})$.
    \item $\boldsymbol{\epsilon}_{\text{CFM}}$ is a \textbf{CFM} residual that treats the unstructured dynamics as a stochastic generative process.
\end{enumerate}


Rather than learning the conservative and residual components independently, we train the full model jointly in an end-to-end manner under a unified objective. 

\section{Methodology}

\subsection{Structured Dynamics Decomposition}

The core of STRIDE is the observation that a robot's motion can be naturally decomposed into structured rigid-body dynamics and residual interaction effects. Starting from the manipulator equation, we express the system acceleration $\ddot{\mathbf{q}}$ as


\begin{equation}
    \ddot{\mathbf{q}} =
    \underbrace{\mathbf{M}^{-1}(\mathbf{q})
    \left[
    \boldsymbol{\tau}
    - \mathbf{C}(\mathbf{q}, \dot{\mathbf{q}})\dot{\mathbf{q}}
    - \mathbf{g}(\mathbf{q})
    \right]}_{f_{\text{struct}}}
    \newline
    + \underbrace{\mathbf{M}^{-1}(\mathbf{q}) \mathbf{F}_{\text{ext}}}_{f_{\text{unstruct}}}.
\end{equation}

Rather than approximating the entire right-hand side with a single black-box model, STRIDE explicitly adopts a dual-stream architecture:

\begin{equation}
\ddot{\mathbf{q}}_{\text{pred}} =
f_{\text{LNN}}(\mathbf{q}, \dot{\mathbf{q}}, \boldsymbol{\tau}; \theta)
+
\mathbf{M}^{-1}(\mathbf{q})
\boldsymbol{\epsilon}_{\text{CFM}}(\mathbf{q}, \dot{\mathbf{q}}, \boldsymbol{\tau}, \mathbf{z}; \phi),
\end{equation}

where,
\[
f_{\text{LNN}} = \mathbf{M}^{-1}(\mathbf{q})
    \left[
    \boldsymbol{\tau}
    - \mathbf{C}(\mathbf{q}, \dot{\mathbf{q}})\dot{\mathbf{q}}
    - \mathbf{g}(\mathbf{q})
    \right],
\]

and $\boldsymbol{\epsilon}_{\text{CFM}}$ represents a stochastic residual ($F_{ext}$) sampled from a generative model with latent noise $\mathbf{z} \sim \mathcal{N}(\mathbf{0}, \mathbf{I})$. $\phi$ and $\theta$ are learnable parameters.

\subsection{The Lagrangian Prior}

To preserve physical structure, STRIDE parameterizes $f_{\text{LNN}}$ using a LNN. Specifically, we learn a network to predict the Cholesky factor $\mathbf{L}_\theta(\mathbf{q})$ and a scalar-valued network to predict the potential energy $\mathcal{V}_\theta(\mathbf{q})$.

\textbf{Positive-Definiteness of the Mass Matrix.}
To ensure physical consistency, STRIDE constructs the mass matrix via a Cholesky factorization
\[
\mathbf{M}_\theta(\mathbf{q}) =
\mathbf{L}_\theta(\mathbf{q}) \mathbf{L}_\theta(\mathbf{q})^\top,
\]
where $\mathbf{L}_\theta$ is lower triangular and its diagonal entries are enforced to be positive using a softplus activation. This guarantees that $\mathbf{M}_\theta(\mathbf{q})$ remains symmetric positive-definite during both training and deployment.

The kinetic energy is then computed as
\[
\mathcal{T}(\mathbf{q}, \dot{\mathbf{q}})
=
\dot{\mathbf{q}}^\top \mathbf{M}_\theta(\mathbf{q}) \dot{\mathbf{q}}.
\]
Together with \eqref{eq:lagrangian} and \eqref{eq:euler_lagrange}, this formulation encourages $f_{\text{LNN}}$ to capture the conservative mechanics of the system, reducing the risk of energy drift commonly observed in purely data-driven models.







\subsection{Generative Residuals via Flow Matching}

The unstructured component $\mathbf{F}_{\text{ext}}$ includes friction, impacts, and other non-conservative interaction effects. Instead of modeling this term deterministically, STRIDE represents it as a conditional stochastic process that captures the distribution of residual forces given the system state.

We employ CFM to learn a continuous transport map from a simple base distribution to the target residual force distribution. Specifically, we sample latent noise $\mathbf{z}_0 \sim \mathcal{N}(\mathbf{0}, \mathbf{I})$ and learn a context-conditioned vector field $v_\phi(\mathbf{z}, t \mid \mathbf{c})$, where the context is $\mathbf{c} = (\mathbf{q}, \dot{\mathbf{q}}, \boldsymbol{\tau})$. The residual force sample is obtained by integrating the learned flow:

\begin{equation}
\boldsymbol{\epsilon}_{\text{CFM}} =
\mathbf{z}_0 + \int_{0}^{1}
v_\phi(\mathbf{z}_t, t \mid \mathbf{c}) \, dt.
\end{equation}


Compared to iterative diffusion sampling, this continuous-time formulation enables efficient generation of physically plausible residual forces (see Sec.~\ref{section:inf_result}), making STRIDE compatible with high-frequency control loops.

\subsection{Joint Optimization}
\label{sec:loss}

A key feature of STRIDE is that the LNN and CFM components are trained jointly under a single supervised objective on observed accelerations. Given a dataset $\mathcal{D} = \{(\mathbf{c}, \ddot{\mathbf{q}})\}$, we minimize

\begin{multline}
\mathcal{J}(\theta, \phi) =
\mathbb{E}_{(\mathbf{c}, \ddot{\mathbf{q}}) \sim \mathcal{D}, \mathbf{z} \sim \mathcal{N}}
\Bigl[
\Bigl\|
\ddot{\mathbf{q}} -
\Bigl(
f_{\text{LNN}}(\mathbf{c}; \theta)
\\
+ \mathbf{M}_\theta^{-1}(\mathbf{q})
\boldsymbol{\epsilon}_{\text{CFM}}(\mathbf{c}, \mathbf{z}; \phi)
\Bigr)
\Bigr\|^2
\Bigr].
\end{multline}

Optimizing both components jointly encourages an implicit division of labor: the LNN captures the low-variance structured dynamics, while the CFM residual models stochastic variability. The overall STRIDE training procedure is summarized in Algorithm~\ref{alg:training}.


\begin{algorithm}[t]
\caption{Joint Training of STRIDE Dynamics}
\label{alg:training}
\KwIn{Dataset $\mathcal{D}$, learning rate $\eta$}
\KwOut{Parameters $\theta, \phi$}
Initialize $\theta, \phi$\;
\While{not converged}{
  Sample batch $(\mathbf{c}, \ddot{\mathbf{q}}) \sim \mathcal{D}$\;
  Compute $f_{\text{LNN}}(\mathbf{c}; \theta)$\;
  Sample $\mathbf{z}_0 \sim \mathcal{N}(\mathbf{0}, \mathbf{I})$\;
  Integrate CFM to obtain $\boldsymbol{\epsilon}_{\text{CFM}}$\;
  Form $\hat{\ddot{\mathbf{q}}}$ and compute loss $\mathcal{L}$\;
  Update $\theta, \phi$\;
}
\end{algorithm}

\subsection{Structural Motivation: Avoiding Averaging Bias}

While the LNN component guarantees the preservation of conservative rigid-body mechanics, representing the unstructured residual $\mathbf{F}_{\text{ext}}$ deterministically introduces a critical failure mode in contact-rich environments as mentioned before. We rely on a well-known property of regression to motivate our generative approach.

Let $p^*(\mathbf{F}_{\text{ext}} \mid \mathbf{c})$ denote the true conditional distribution of non-conservative interaction forces given the context $\mathbf{c} = (\mathbf{q}, \dot{\mathbf{q}}, \boldsymbol{\tau})$. In many robotic interaction scenarios involving intermittent contact or frictional regime changes, this distribution can become inherently multi-modal, with support concentrated on a set of physically realizable modes $\mathcal{M}$.

A common modeling strategy is to approximate the residual forces using a deterministic regressor trained with a mean-squared error objective. Under the standard $L_2$ risk,
\begin{equation}
\min_{f} \; \mathbb{E}\left[\left\| \mathbf{F}_{\text{ext}} - f(\mathbf{c}) \right\|^2\right],
\end{equation}
the optimal predictor is the conditional mean
\begin{equation}
f^*(\mathbf{c}) = \mathbb{E}_{p^*}[\mathbf{F}_{\text{ext}} \mid \mathbf{c}].
\end{equation}

When the true conditional distribution is multi-modal, this mean may lie between physically distinct modes. In robotic systems, such averaging can smooth over discontinuous interaction phenomena, for example, when a foot either slips or establishes stiction depending on subtle state variations. As a result, deterministic residual models may produce force predictions that do not correspond to any physically realizable interaction outcome.

To address this limitation, we model the residual forces using a conditional generative formulation. Specifically, CFM learns a transport vector field that maps simple noise to samples from the conditional distribution,
\begin{equation}
\mathbf{F}_{\text{ext}} \sim p_{\phi}(\mathbf{F}_{\text{ext}} \mid \mathbf{c}, \mathbf{z}),
\end{equation}
enabling the model to represent multi-modal interaction behavior. 


\textbf{Practical implication.} The generative residual formulation is particularly beneficial in regimes where interaction dynamics exhibit branching behavior, such as near contact transitions or frictional regime changes. In these settings, sampling-based prediction with CFM provides a more faithful representation of the underlying uncertainty while remaining compatible with model-based control.

\subsection{Discussion on Residual Allocation}

A potential concern is whether the residual component may absorb dynamics that should be captured by the structured model. In STRIDE, the LNN is constrained by the Euler–Lagrange formulation and a positive-definite mass matrix, which biases it toward modeling conservative dynamics. Empirically, in a validated bouncing pendulum experiment comparing analytical priors with non-parametric residuals, we observe a clear separation of responsibilities: during smooth rigid-body motion (e.g., flight or free swing), the generative residual contributes only 6.5\% of the predicted acceleration magnitude. In contrast, during contact transitions and boundary impacts—where structured Lagrangian assumptions degrade—the residual dynamically scales to 78.8\%, absorbing high-frequency discontinuities without corrupting the global mass matrix.


\section{Experiments and Results}

We evaluate whether the proposed STRIDE formulation delivers measurable gains in prediction accuracy, control performance, and hardware deployment. Specifically, we assess: (1) long-horizon predictive accuracy and stability under compounding rollout errors, (2) accuracy in modeling non-conservative effects, including contact impulses and impact-induced discontinuities, and (3) real-time deployability within existing control pipelines on hardware. Across both the Unitree Go1 quadruped and the Unitree G1 humanoid, we demonstrate consistent improvements in long-horizon rollout stability and contact-force prediction, validating the effectiveness of the proposed model.

We further examine the broader implications of this modeling approach by analyzing its behavior in systems where sensitivity to bias, such as systems with unstable equilibria, can help evaluate model bias.


\subsection{Baselines}
We compare STRIDE against deterministic and generative baselines trained under controlled and identical settings, including same network size and training iterations. All models are evaluated using identical rollouts to ensure a fair comparison.

\begin{itemize}

\item \textbf{ONN Dynamics (Black-Box MLP):}  
A fully data-driven baseline that directly predicts the next environment observation.  
\textit{Input:} Current observation $\mathbf{o}_t$.  
\textit{Output:} Predicted next observation $\hat{\mathbf{o}}_{t+1}$.  
\textit{Loss:} RMSE between predicted and ground-truth next observations.  
\textit{Role:} Serves as an unstructured regression baseline without explicit physical inductive bias.

\item \textbf{DeLaN:}  
A physics-informed baseline based on Deep Lagrangian Networks that parameterizes kinetic and potential energy to enforce Euler Lagrange structure \cite{lutter2019deep}.  
\textit{Input:} $(\mathbf{q}, \dot{\mathbf{q}}, \boldsymbol{\tau})$.  
\textit{Output:} Predicted joint acceleration $\hat{\ddot{\mathbf{q}}}$.  
\textit{Loss:} RMSE on predicted accelerations.   
\textit{Role:} Tests the benefit of structured dynamics without stochastic residual modeling.

\item \textbf{Pure Diffusion Dynamics:}  
A fully generative baseline that models the conditional distribution of the next observation using a diffusion model.  
\textit{Input:} Current observation $\mathbf{o}_t$ and diffusion timestep.  
\textit{Output:} Sampled next observation $\hat{\mathbf{o}}_{t+1}$.  
\textit{Loss:} Standard diffusion denoising objective with RMSE evaluation in observation space.  
\textit{Role:} Evaluates fully generative modeling without explicit physical structure.

\item \textbf{LNN + Diffusion:}  
A baseline in which an LNN models nominal rigid-body dynamics and a diffusion model predicts a residual acceleration.  
\textit{Input:} $(\mathbf{q}, \dot{\mathbf{q}}, \boldsymbol{\tau})$ and diffusion noise.  
\textit{Output:} Predicted acceleration  
\[
\hat{\ddot{\mathbf{q}}} = f_{\text{LNN}} + \mathbf{M}^{-1}(\mathbf{q})\,\epsilon_{\text{diff}}.
\]  
\textit{Loss:} RMSE on total predicted acceleration with diffusion denoising objective for the residual.   
\textit{Role:} Tests whether flow-based residuals in STRIDE provide advantages over diffusion within the same structured setting.

\item \textbf{STRIDE (Ours):}  
STRIDE combines a Lagrangian Neural Network prior with a CFM residual. The model is trained end-to-end using the joint objective described in Sec.~\ref{sec:loss}.  
\textit{Input:} $(\mathbf{q}, \dot{\mathbf{q}}, \boldsymbol{\tau})$ and latent noise $\mathbf{z}$.  
\textit{Output:} Predicted joint acceleration.
\textit{Loss:} Joint RMSE-based objective on accelerations with flow-matching training for the residual. 
\textit{Role:} Evaluates the proposed structured–stochastic decomposition.

\end{itemize}

\subsection{MPC Integration and Deployment}

To evaluate the practical utility of the learned dynamics model, we embed STRIDE within a sampling-based MPC framework, specifically Model Predictive Path Integral (MPPI) control. MPPI is well suited to this setting because it performs optimization via sampling and naturally accommodates stochastic dynamics without requiring analytic gradients or local linearization \cite{shirwatkar2025pip}. 


We integrate STRIDE into the Dreamer–MPC control architecture of \cite{kotecha2025realtime}, where reinforcement learning is used to learn a warm-start policy along with dynamics, reward, and value models, while MPPI performs online trajectory optimization at deployment. This setup provides a stringent evaluation: the receding-horizon MPC loop repeatedly exposes long-horizon prediction errors, the modular design enables controlled swapping of dynamics models for fair comparison, and the contact-rich nature of legged locomotion stresses modeling assumptions.

To further evaluate adaptability, we extend the study to a variable gait optimization framework for quadrupeds based on the same Dreamer–MPC backbone \cite{kotecha2025investigating}. In this formulation, MPPI jointly optimizes actions and continuous gait parameters online, enabling seamless transitions between locomotion modes without explicitly training each transition in reinforcement learning. This provides a stronger test of predictive quality, as the planner explores previously unseen gait configurations and relies heavily on accurate short-horizon rollouts. We apply this formulation to the quadruped platform in both simulation and hardware experiments.

\subsection{Long-Horizon Prediction and Stability}
Long-horizon stability is a primary motivation for incorporating a structured Lagrangian prior, as physically consistent inductive bias can mitigate the exponential error growth typically observed in unstructured regressors.

\begin{figure}
    \centering
    \includegraphics[width=0.9\linewidth]{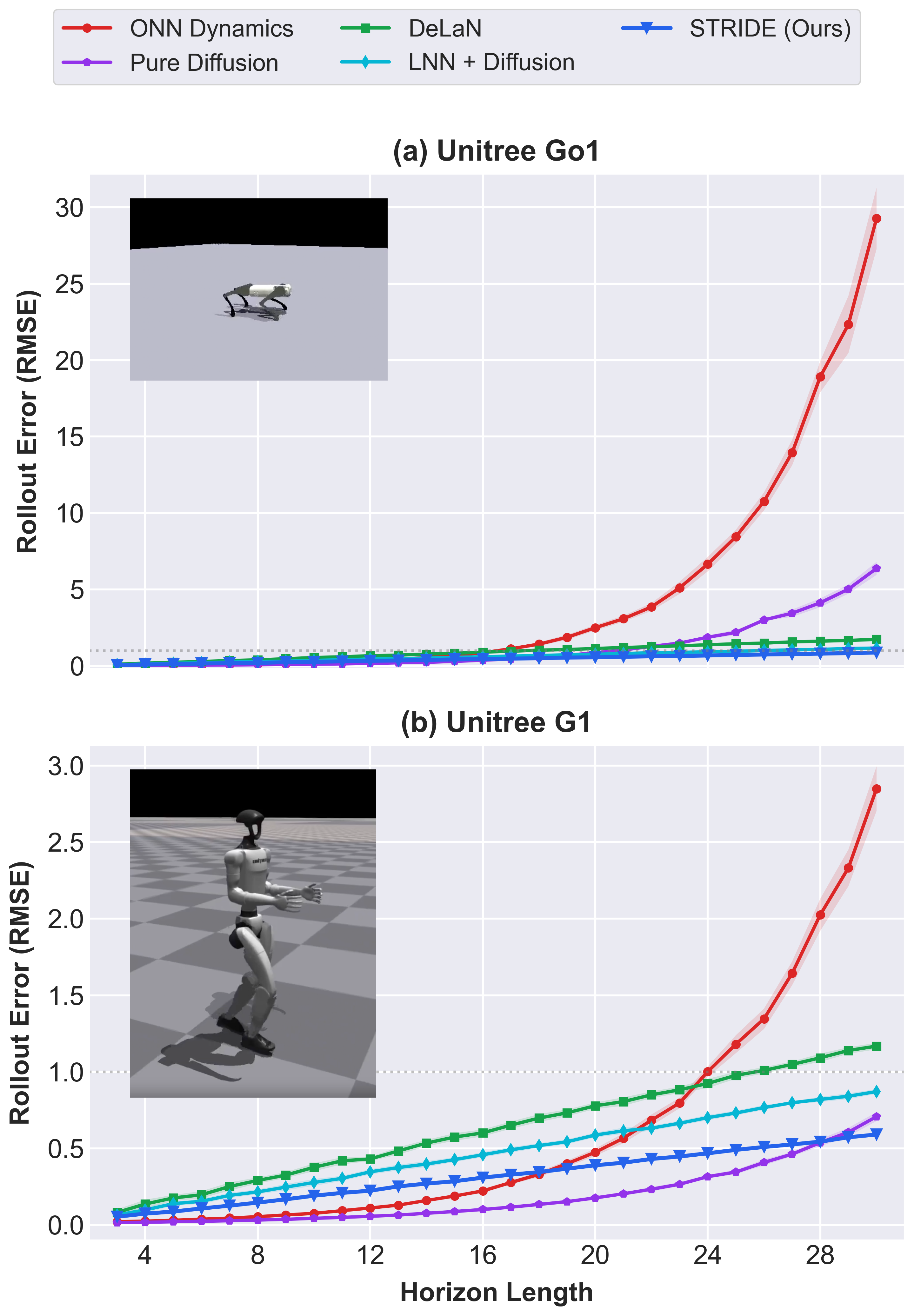}
    \caption{Long-horizon rollout error on complex legged systems (Unitree Go1 and Unitree G1). The ONN baseline exhibits rapid, near-exponential error growth, while the DeLaN reduces error growth to approximately linear. STRIDE further reduces long-horizon drift by capturing stochastic, contact-induced variability. \vspace{-10pt}}
    \label{fig:rollout error}
\end{figure}

We evaluate long-horizon predictive accuracy via multi-step rollouts for both Unitree Go1 and Unitree G1 robots in simulation and report the cumulative root-mean-square error (RMSE) over a horizon of $H=30$ (Fig.~\ref{fig:rollout error}). As expected, ONN exhibits a rapid (effectively exponential) increase in error even within 30 time steps. The DeLaN baseline substantially improves stability, showing a slower, approximately linear growth in error consistent with the benefits of enforcing structure in the conservative dynamics. Augmenting the LNN with generative residuals further reduces drift and improves long-horizon prediction on both Unitree Go1 and Unitree G1.

Table~\ref{tab:hardware_results} reports the cumulative normalized RMSE over the 30-step horizon. STRIDE reduces rollout error by 83\% relative to the ONN baseline on Unitree Go1 (53\% on Unitree G1), and achieves a further 19\% (Unitree Go1) and 21\% (Unitree G1) reduction compared to the strongest structured generative baseline (LNN + Diffusion), representing a substantial improvement over models that already incorporate physical priors. Interestingly, the Pure Diffusion baseline performs slightly better on the Unitree G1 humanoid, likely because direct observation-space modeling can more readily absorb high-dimensional sensing noise, whereas structure-constrained models prioritize physically consistent acceleration prediction.

\begin{figure}
    \centering
    \includegraphics[width=\linewidth]{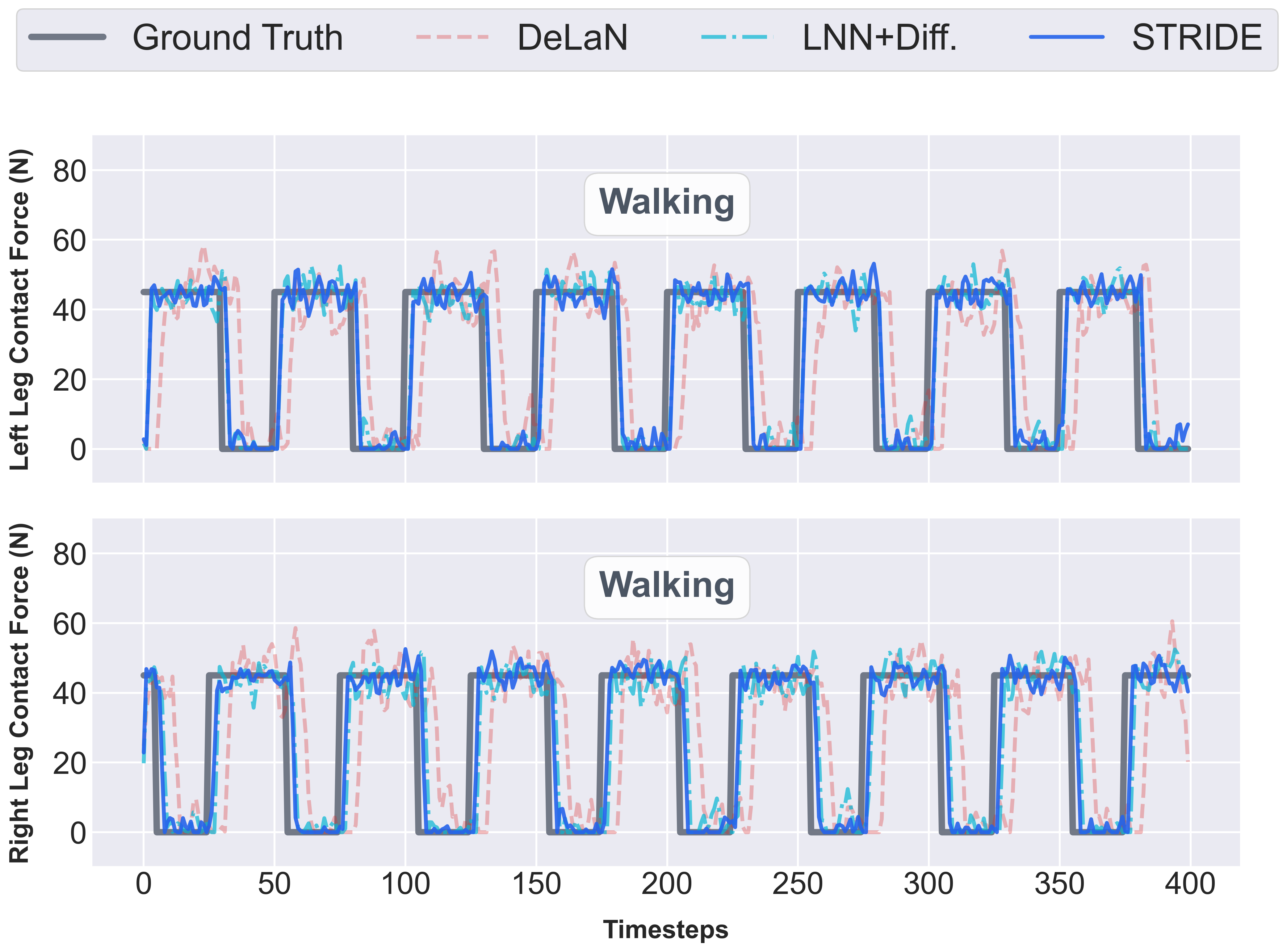}
    \caption{Contact force prediction on the Unitree G1 humanoid during walking. Predicted vertical ground reaction forces are compared against ground truth for left and right legs. STRIDE closely tracks the timing and magnitude of stance–swing transitions, preserving sharp impact discontinuities and reducing force smoothing observed in deterministic baselines (DeLaN and LNN+Diff). }
    \label{fig:contact_G1} 
\end{figure}

\begin{figure}
    \centering
    \includegraphics[width=\linewidth]{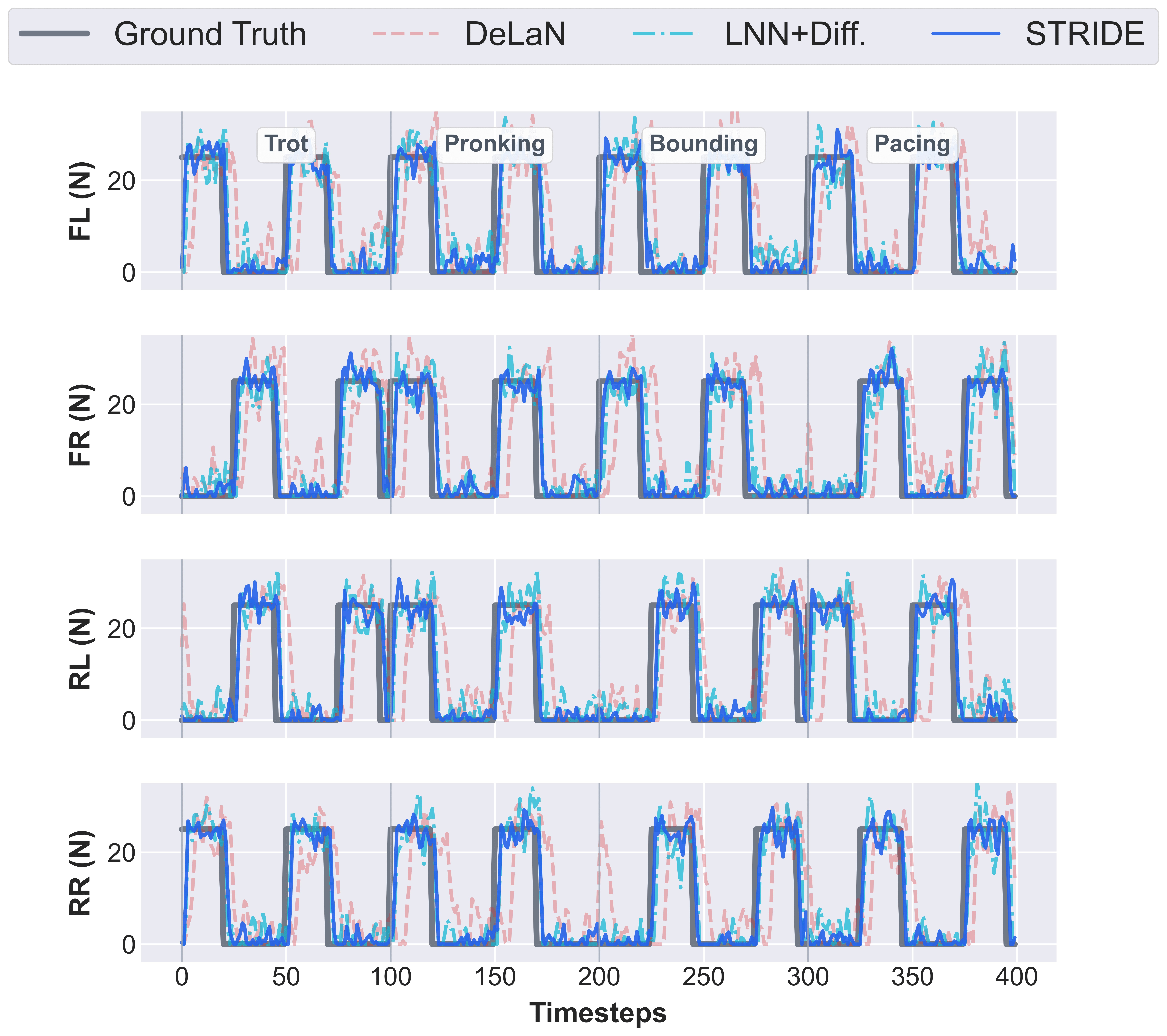}
    \caption{Contact force prediction on the Unitree Go1 quadruped across multiple gaits (trot, pronk, bound, and pace). Vertical ground reaction forces are shown for all four legs (FL, FR, RL, RR). \vspace{-10pt}}
    \label{fig:contact_go1} 
\end{figure}

\subsection{Contact and Force Prediction}

We next evaluate contact modeling accuracy by comparing predicted ground reaction forces against ground-truth measurements, as shown in Fig.~\ref{fig:contact_G1} and Fig.~\ref{fig:contact_go1}. State RMSE in Table~\ref{tab:hardware_results} corresponds to multi-step open-loop prediction error in observation space evaluated over a 30-step horizon. Since the plotted contact forces correspond to the non-conservative term $F_{\text{ext}}$, they are directly available only for LNN-based models; therefore, force-level comparisons are limited to structured baselines.

Beyond reducing aggregate RMSE, STRIDE captures the sharp discontinuities at impact and during swing–stance transitions, preserving both the timing and magnitude of impulsive contact forces, as illustrated in Fig.~\ref{fig:contact_go1}. The achieved force prediction error corresponds to approximately 13\% of the nominal contact force magnitude on Unitree Go1 and 8\% on Unitree G1, indicating accurate modeling relative to the physical scale of interaction forces. Improvements are consistent across individual legs and across the four evaluated gaits on the quadruped, suggesting that the model generalizes across contact configurations rather than overfitting to a specific locomotion mode.

Overall, the proposed model achieves substantially lower force prediction error than all structured baselines, as summarized in Table~\ref{tab:hardware_results}. STRIDE reduces contact force prediction error by approximately 30\% relative to the DeLaN baseline on both quadruped and humanoid experiments. This reduction contributes to more reliable contact-aware planning and improved closed-loop stability during hardware deployment. The consistent error reduction observed on the higher-dimensional humanoid system further indicates that STRIDE generalizes and scales robustly to more complex platforms.


\begin{figure}[!h]
    \centering
    \includegraphics[width=0.9\linewidth]{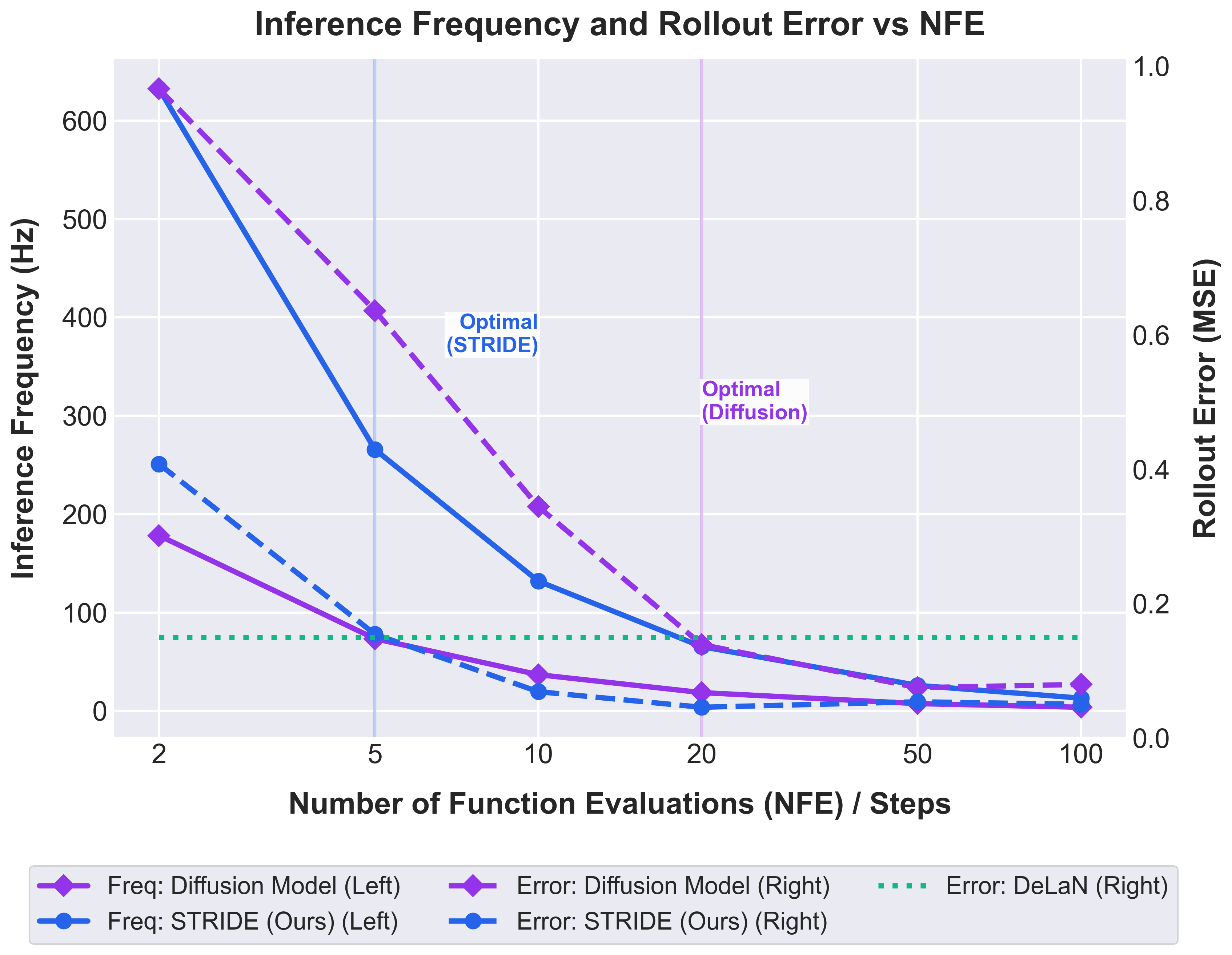}
    \caption{Inference-time comparison between diffusion sampling and CFM. Dotted curves show rollout error versus NFEs, and solid curves show inference frequency. CFM attains DeLaN-level rollout error with fewer evaluations while sustaining higher sampling frequency, highlighting its suitability for real-time deployment.}
    \label{fig:inf}
\end{figure}

\begin{figure*}
    \centering
    \includegraphics[width=\linewidth]{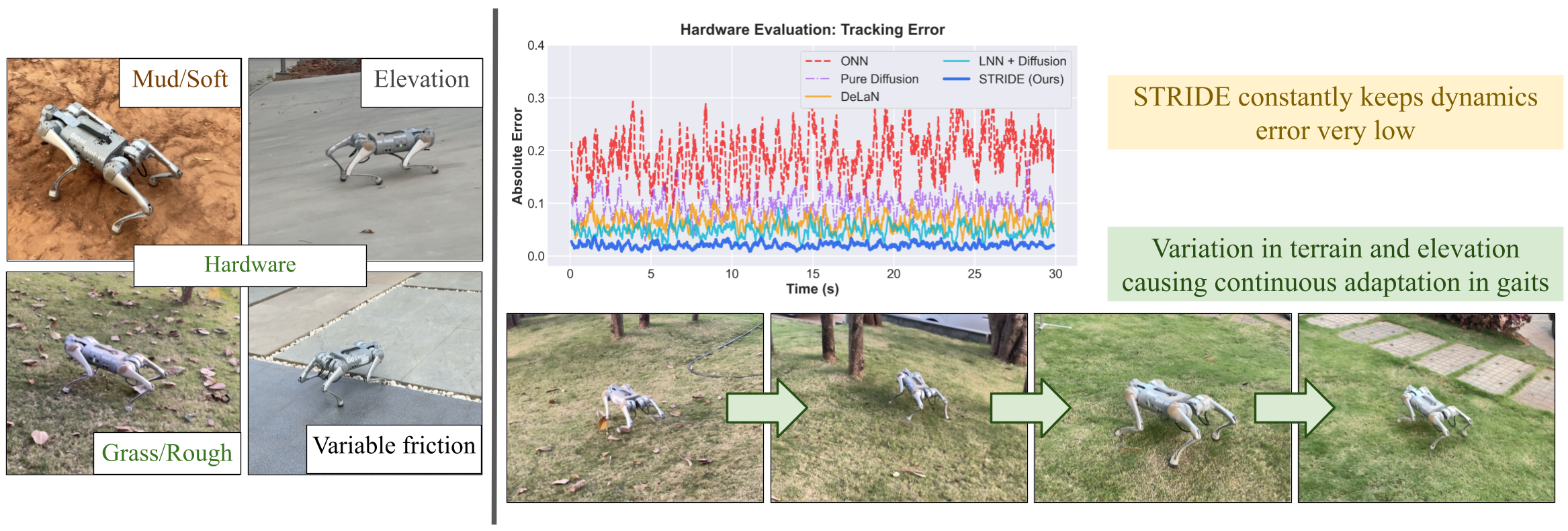}
    \caption{Hardware deployment of STRIDE within the Dreamer–MPPI multi-gait framework on the Unitree Go1. STRIDE maintains consistently low tracking error under sensing noise and hardware uncertainties while enabling smooth gait transitions across varying speeds and elevations. The controller demonstrates zero-shot adaptation to unseen terrains, including friction changes, 20° slopes, muddy, and grassy surfaces.}
    \label{fig:hardware_img}
\end{figure*}

\begin{figure*}
    \centering
    \includegraphics[width=\linewidth]{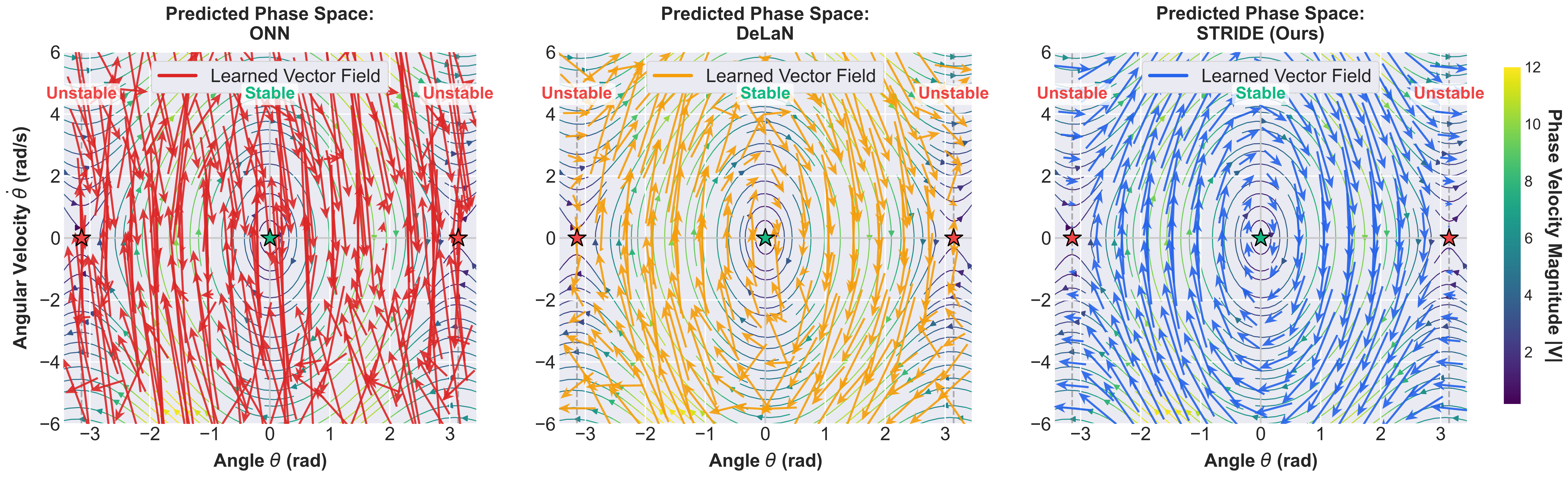}
    \caption{Phase portrait on the 1-DoF pendulum near the unstable upright equilibrium. STRIDE preserves physically consistent trajectories and captures multi-modal next-state evolution, while deterministic baselines (e.g., DeLaN) show averaging bias and drift in sensitive regions. \vspace{-10pt}}
    \label{fig:phase}
\end{figure*}

\begin{table}[t]
\centering
\caption{Performance comparison. (Evaluated at Horizon = 30) \vspace{-5pt}}
\label{tab:hardware_results}
\begin{tabular}{llcc}
\toprule
\textbf{Robot} & \textbf{Model} & \textbf{State RMSE} $\downarrow$ & \textbf{Force Error (N)} $\downarrow$ \\
\midrule
\multirow{5}{*}{Go1}
& Plain MLP & 5.489 & NA \\
& Pure Diffusion & 3.521 & NA \\
& DeLaN & 1.326 & 9.4 \\
& LNN + Diffusion & 1.154 & 8.8 \\
& \textbf{STRIDE (Ours)} & \textbf{0.932} & \textbf{6.7} \\
\midrule
\multirow{5}{*}{G1}
& Plain MLP & 1.204 & NA \\
& Pure Diffusion & 0.314 & NA \\
& DeLaN & 1.023 & 17.1 \\
& LNN + Diffusion & 0.918 & 15.4 \\
& \textbf{STRIDE (Ours)} & \textbf{0.289} & \textbf{12.1} \\
\bottomrule
\vspace{-20pt}
\end{tabular}
\end{table}

\subsection{Inference and Accuracy Comparison} \label{section:inf_result}
Figure~\ref{fig:inf} compares STRIDE’s CFM-based residual generation with the diffusion residual used in the LNN + Diffusion baseline, evaluating both computational cost and rollout accuracy. As shown in Fig.~\ref{fig:inf}, CFM achieves substantially higher inference frequency across all evaluated numbers of function evaluations (NFEs). The advantage is particularly pronounced at low NFE budgets, highlighting the efficiency of continuous-time flow transport relative to iterative denoising procedures.

The figure further shows that the flow-based model reaches the rollout error level of the DeLaN baseline with significantly fewer NFEs, whereas diffusion requires more evaluations to attain comparable accuracy. Overall, STRIDE exhibits a favorable accuracy–efficiency trade-off. These results justify the choice of CFM within STRIDE for real-time model-based control, where both predictive accuracy and sampling speed are essential.

\subsection{Hardware Implementation and Real-Time Performance}

To thoroughly test the dynamics model beyond simulation, we implement the Dreamer–MPC multi-gait controller on hardware using the Unitree Go1. Real-world deployment provides a more stringent evaluation due to unmodeled compliance, actuator dynamics, sensing noise, terrain variation, slipping, and other uncertainties.


We evaluated STRIDE-embedded control pipeline under varying commanded speeds and terrain elevations, where the system demonstrates smooth and continuous gait adaptation for stable locomotion. The improved dynamics model enables accurate long-horizon rollouts, allowing MPPI to effectively optimize over gait parameters and control actions in real time. As the hardware platform does not provide direct measurements of contact forces, we instead evaluate performance through rollout error comparisons. As shown in Fig.~\ref{fig:hardware_img}, STRIDE maintains the lowest prediction error in the presence of sensing noise and hardware uncertainties.

We further evaluate zero-shot adaptation across four previously unseen terrains, including transitions from high-friction to low-friction surfaces, 20° slopes, as well as muddy and grassy terrain. The controller successfully adapts gaits and joint positions under these conditions, maintaining stable locomotion without additional training or parameter tuning. The model captures shifts in residual non-conservative forces ($\mathbf{F}_{\text{ext}}$), enabling the MPPI optimizer to adjust gait and refine actions online in response to changing contact dynamics. STRIDE runs in 3 ms within the MPPI loop, maintaining an on-board control frequency of 50 Hz, thereby demonstrating real-time feasibility alongside improved robustness. 

In summary, our experiments successfully demonstrate:

\begin{itemize}
    \item Velocity tracking from 0 to 2 m/s, including sudden velocity change.
    \item Adaptation to changing elevation up to 20°.
    \item Single-shot adaptation to 4 previously unseen terrains: 
    \begin{itemize}
        \item Muddy: increased compliance of ground
        \item Grassy: increased roughness and disturbances
        \item Terrain with high and low friction surface
        \item Grassy and flat terrain with increasing slope 
    \end{itemize}
    \item Enables smooth and continuous gait adaptation, including various gaits like trotting, pronking and pacing.
\end{itemize}


\subsection{Behavior in Sensitive Dynamical Regimes}

While the primary evaluation focuses on contact-rich legged systems, we further analyze the behavior of the proposed model in a controlled low-dimensional setting to better understand its qualitative properties. Specifically, we study a 1-DoF simple pendulum system to examine the consistency of phase portraits in sensitive dynamical regions.

Despite its simplicity, the pendulum exhibits a stable configuration and an unstable equilibrium at the upright configuration. Trajectories near this equilibrium are highly sensitive to modeling errors and stochastic perturbations, making it an informative regime for evaluating bias and uncertainty representation. We therefore analyze the predicted phase portrait near the unstable equilibrium (Fig.~\ref{fig:phase}).

In this sensitive region, deterministic predictors such as the ONN and DeLaN exhibit clear distortions in the learned vector field. The ONN produces irregular and noisy flows, while the DeLaN, although structurally smoother, shows noticeable deviations near the unstable equilibrium. In contrast, STRIDE produces a coherent phase portrait that closely aligns with the ground-truth flow. The elliptical orbits around the stable equilibrium are preserved, and the saddle-like structure near the unstable upright configuration remains well defined. These observations indicate that explicitly modeling stochastic residual dynamics mitigates data-induced averaging effects while preserving the qualitative topology of the underlying system.
\vspace{-9pt}
\section{Conclusion}

We presented STRIDE, a dynamics modeling framework that decomposes robot dynamics into conservative rigid-body mechanics and stochastic non-conservative interaction effects. The conservative component is modeled using a Lagrangian Neural Network to preserve energy-consistent structure and positive-definite inertial properties, while residual forces are represented via Conditional Flow Matching (CFM) to capture multi-modal contact variability with efficient sampling. Across complex systems, including the Unitree Go1 quadruped and Unitree G1 humanoid, STRIDE achieves at least a 20\% improvement in long-horizon prediction accuracy and contact-force modeling compared to deterministic and diffusion-based baselines, with these gains translating to stable hardware behavior. The flow-based residual offers a favorable accuracy–efficiency trade-off, requiring fewer function evaluations than diffusion-based alternatives. Pendulum experiments further demonstrate improved preservation of phase-space structure near unstable equilibria, mitigating averaging effects observed in deterministic predictors.

Future work will explore tighter integration with uncertainty-aware and risk-sensitive planning and improved robustness under distribution shift. Another promising direction is to extend the flow-matching residual to incorporate visual observations, enabling perception-driven residual modeling that can capture environment-dependent interaction effects within the same physics-informed framework.

\vspace{-5pt}
\bibliographystyle{ieeetr}
\bibliography{references}

\end{document}